%% file: VECTAR2010sixpoints.tex
\documentclass[runningheads]{llncs}
\usepackage{graphicx}
\usepackage{amsmath,amssymb} 
\usepackage{lineno}
\usepackage{color}
\usepackage[lined,algonl,boxed]{algorithm2e}

\input{bold}

\input{blackboardbold}

\begin{document}

\renewcommand\thelinenumber{
    \color[rgb]{0.2,0.5,0.8}\normalfont\sffamily\scriptsize\arabic{linenumber}\color[rgb]{0,0,0}
    }
\newcommand{\point}{
    \raise0.7ex\hbox{.}
    }


\pagestyle{headings}

\mainmatter

\title{Sparse motion segmentation using Multiple
Six-Point Consistencies
\thanks{This work has been supported by ELLITT, the strateigc area for ICI research funded by the Swedish goverment.}
} 

\titlerunning{Sparse motion segmentation using Multiple
Six-Point Consistencies}

\authorrunning{V. Zografos et al.}

\author{Vasileios Zografos
, Klas Nordberg and Liam Ellis}
\institute{Computer Vision Laboratory, Link\"{o}ping University, Sweden \\
\email{ \{zografos,klas,liam\}@isy.liu.se}
}

\maketitle

\begin{abstract}
We present a method for segmenting an arbitrary number of moving
objects in image sequences using the geometry of 6 points in 2D to infer motion
consistency.  The method has been evaluated on the Hopkins~155 database
and surpasses current state-of-the-art methods such as SSC, both in
terms of overall performance on two and three motions but also in terms of maximum errors.  The method works by finding initial
clusters in the spatial domain, and then classifying each
remaining point as belonging to the cluster that minimizes a motion
consistency score.  In contrast to
most other motion segmentation methods that are based on an affine
camera model, the proposed method is fully projective.
\end{abstract}

\input{introduction}

\input{background}
\input{algorithm}
\input{experiments}
\input{conclusion}

\bibliographystyle{splncs}
\bibliography{VECTAR2010sixpoints}

\end{document}

%% file: bold.tex
\def\bl{{\bf l}}

\def\bs{{\bf s}}

\def\bx{{\bf x}}
\def\by{{\bf y}}
\def\bz{{\bf z}}

\def\bA{{\bf A}}

\def\bC{{\bf C}}

\def\bH{{\bf H}}

\def\bL{{\bf L}}

\def\bT{{\bf T}}

\def\bZ{{\bf Z}}



%% file: blackboardbold.tex
\DeclareSymbolFont{AMSb}{U}{msb}{m}{n}

\DeclareMathSymbol{\bbP}{\mathbin}{AMSb}{"50}
\DeclareMathSymbol{\bbQ}{\mathbin}{AMSb}{"51}
\DeclareMathSymbol{\bbR}{\mathbin}{AMSb}{"52}

%% file: introduction.tex
\section{Introduction}

Motion segmentation can be defined as the task of separating a sequence
of images into different regions, each corresponding to a distinct
rigid motion. There are several strategies for solving the motion
segmentation problem, some of which are based on first producing a
dense motion field, using optical flow techniques, and then analyzing
this field.  Examples of this approach are \cite{BlackJepson1996PAMI}
where the optic flow is given as a parametric model and the parameters are
determined for each distinct object, or the normalised graph cuts by
\cite{ShiMalikPAMI2000}. 

Other approaches are instead applied to a
sparse set of points, typically interest points that are tracked over
time, and their trajectories analysed in the image. 
A common simplifying assumption is that only small depth variations occur and an
affine camera model may be used.  The problem can then be solved using
the factorization method by \cite{TomasiKanade1992IJCV}. This approach has attracted a large interest in recent literature, with the
two current state-of-the-art methods, relative to standard datasets such as Hopkins~155
\cite{TronVidal2007CVPR}, being Sparse Subspace Clustering (SSC)
\cite{ElhamifarVidal2009CVPR} and Spectral Clustering of linear
subspaces (SC) \cite{LauerSchnorrICCV2009}. 

Other common methods in the literature are based on Spectral Curvature
Clustering (SCC) \cite{ChenLerman2009ICCV}, penalised MAP estimation of
mixtures of subspaces using linear programming (LP) \cite{HuBMVC2009},
Normalised Subspace Inclusion (NSI) \cite{SilvaICCV2009}, Non-negative
Matrix Factorisation (NNMF) \cite{CheriyadatICCV2009}, Multi-Stage
unsupervised Learning (MSL) \cite{SugayaKanatani2004SMVP}, Local
Subspace Affinity (LSA), Connected Component Search (CCS)
\cite{Roweis2003}, unsupervised manifold clustering using LLE (LLMC)
\cite{GohCVPR2007}, Agglomerative Lossy
Compression (ALC) \cite{RaoEtAl2008CVPR}, Generalised Principal
Component Analysis (GPCA) \cite{VidalEtAlIJCV2008}, or on RANdom SAmple
Consensus (RANSAC) \cite{TronVidal2007CVPR}.

In this paper we describe a motion segmentation method for sparse point
trajectories, which is based on the previous work on six point consistency (SPC) \cite{NordZogICPR2010}, but
with the additional novelties and improvements: (i) an alternative method
for estimating the vector $\bs$ (Sec.~\ref{sec:ests}), (ii) a new
matching score (Sec.~\ref{sec:score}), and (iii) a modified
classification algorithm (Sec.~\ref{sec:algo}).

%% file: background.tex
\vspace*{-1mm}
\section{Mathematical background}
\label{sec:mback}

Our proposed method uses the consistent motion in
the image plane generated by 6 points located on a rigid 3D object. The
mathematical foundation of this theory was formulated by Quan
\cite{Quan1995PAMI} and later extended by other authors
\cite{Carlsson1995WRVS,WeinshallWermanShashua1996ECCV,TorrZisserman1997IVC}.
A similar idea was presented in \cite{Nordberg2007ISVC}, and later used
for motion segmentation in \cite{NordZogICPR2010}. \cite{Nordberg2007ISVC} shows
that the consistency test can be formulated as a constraint directly on
the image coordinates of the 6 points and that, similarly to epipolar
lines emerging from the epipolar constraint, this 6-point constraint
generates 6 lines that each must intersect its corresponding point.

More formally, we consider a set of six 3D points, with homogeneous coordinates
$\bx_{k}$, projected onto an image according to the pinhole camera
model:
\begin{equation}
  \by_{k} \sim \bC \: \bT \: \bx_{k}, \quad k = 1, \ldots, 6,
  \label{eq:ykisCHxk}
\end{equation}
where $\by_{k}$ are the corresponding homogeneous image coordinates,
$\bC$ is the 3$\times$4 camera matrix, and $\sim$ denotes equality up
to a scalar multiplication.  $\bT$ is a $4 \times 4$ time dependent
transformation matrix that rotates and translates the set of 3D points
from some reference configuration to the specific observation that
produces $\by_{k}$.  This implies that also $\by_{k}$ is time
dependent. The problem addressed here is how we can determine if an
observed set of image points $\by_{k}$ really is given by
\eqref{eq:ykisCHxk} for a particular set of 3D points $\bx_{k}$ but
with $\bC$ and $\bT$ unknown.

In general, the homogeneous coordinates of the 3D points can be
transformed by a suitable 3D homography $\bH_{\bx}$ to
\textit{canonical} homogeneous 3D coordinates $\bx'$=$\bH_{\bx} \:
\bx$, and similarly, for a particular observation of the image points
we can transform them to canonical homogeneous 2D coordinates
$\by'_{k}$=$\bH_{\by} \: \by_{k}$.  The canonical coordinates are given
by:
\begin{equation}
  ( \bx'_{1} \: \bx'_{2} \: \bx'_{3} \: \bx'_{4} \: \bx'_{5} \: \bx'_{6} )
  \! \sim\sim \!
  \begin{pmatrix}
    1 & 0 & 0 & 0 & 1 & X \\
    0 & 1 & 0 & 0 & 1 & Y \\
    0 & 0 & 1 & 0 & 1 & Z \\
    0 & 0 & 0 & 1 & 1 & T
  \end{pmatrix},
  ( \by'_{1} \:\: \by'_{2} \:\: \by'_{3} \:\: \by'_{4} \:\: \by'_{5} \:\: \by'_{6} )
  \! \sim\sim \!
  \begin{pmatrix}
    1 & 0 & 0 & 1 & u_{5} & u_{6} \\
    0 & 1 & 0 & 1 & v_{5} & v_{6} \\
    0 & 0 & 1 & 1 & w_{5} & w_{6}
  \end{pmatrix}.
  \notag
\end{equation}
Here $\sim\sim$ denotes equality up to an individual scalar
multiplication on each column.  $\bH_{\bx}$ and $\bH_{\by}$ depend on
the 3D points $\bx_{1}$,...,$\bx_{5}$ and on the image points
$\by_{1}$,...,$\by_{4}$, respectively, and after these transformation
are made the relation between 3D points and image points is given by
$\by'_{k}$$\sim$$\bH_{\by} \: \bC \: \bT \: \bH_{\bx}^{-1} \:
\bx'_{k}.$
\newcommand{\tI}{{\tilde I}}
One of the main results in \cite{Quan1995PAMI} is that from these
transformed coordinates we can compute a set of five \textit{relative
invariants} of the image points, denoted $i_{k}$, and of the 3D points,
denoted $\tI_{k}$, according to:
\begin{equation}
  \bz  =   \begin{pmatrix}
     i_{1}  \\
     i_{2}  \\
     i_{3}  \\
     i_{4}  \\
     i_{5}
  \end{pmatrix}   =
  \begin{pmatrix}
     w_{6} (u_{5}  -  v_{5})  \\
     v_{6} (w_{5}  -  u_{5})  \\
     u_{5} (v_{6}  -  w_{6})  \\
     u_{6} (v_{5}  -  w_{5})  \\
     v_{5} (w_{6}  -  u_{6})
  \end{pmatrix},
  \quad \quad
  \bs  =  \begin{pmatrix}
     \tI_{1}  \\
     \tI_{2}  \\
     \tI_{3}  \\
     \tI_{4}  \\
     \tI_{5}
  \end{pmatrix}  =
  \begin{pmatrix}
     X Y  -  Z T  \\
     X Z  -  Z T  \\
     X T  -  Z T  \\
     Y Z  -  Z T  \\
     Y T  -  Z T
  \end{pmatrix}
  \label{eq:defs}
\end{equation}
such that they satisfy the constraint
$\bz \cdot \bs = i_{1} \: \tI_{1} + i_{2} \: \tI_{2} + i_{3} \: \tI_{3} +  i_{4} \: \tI_{4} + i_{5} \: \tI_{5} = 0.$

To realize what this means, we notice that this constraint includes
scalars derived from the reference 3D coordinates $\bx_{k}$ (before
they are transformed) and observed image points $\by_{k}$ (after the
transformation $\bT$ is made), but neither $\bC$ nor $\bT$ are
explicitly included. Therefore, the constraint is satisfied regardless
of how we transform the 3D points (or move the camera), as long as they
are all transformed by the same $\bT$. As long as the observed image
coordinates are consistent with \eqref{eq:ykisCHxk}, the corresponding
relative image invariants $\bz$ must satisfy the constraint for
a fixed $\bs$ computed from the 3D reference points. The canonical
transformations $\bH_{\bx}$ and $\bH_{\by}$ can conveniently be
included into the unknowns $\bC$ and $\bT$.
In short, the above constraint is necessary but not sufficient for the matching between the observed image points and the
3D reference points.

\vspace*{-1mm}
\subsection{The 6-point matching constraint}
\label{sec:6pmc}

The matching constraint is expressed in terms
of the relative invariants $\bz$ and $\bs$ that have been derived by
transforming image and 3D coordinates.  In particular, this means that
it cannot be applied directly onto the image coordinates, similar to
the epipolar constraint.  The transformation $\bH_{\by}$ is
\textit{not} a linear transformation on the homogeneous image
coordinates since it also depends on these coordinates (see the
Appendix of \cite{Quan1995PAMI}).  If however, we make an explicit
derivation of how $\bz$ depends on the 6 image points, it turns out that
it has a relatively simply and also useful form:
\begin{equation}
  \bz = \alpha \begin{pmatrix}
    D_{126} D_{354} \\ D_{136} D_{245} \\ D_{146} D_{253} \\ D_{145} D_{263} \\ D_{135} D_{246}
  \end{pmatrix},
  \quad \quad
  \begin{array}{c}
  \alpha = \dfrac{D_{123}}{D_{124} D_{234} D_{314}}, \\[8mm]
  D_{ijk} = (\by_{i} \times \by_{j}) \cdot \by_{k} =
  \det \begin{pmatrix} \by_{i} & \by_{j} & \by_{k} \end{pmatrix}.
  \end{array}
  \label{eq:zisDD}
\end{equation}
Since $\bz$ can be represented as a projective element, the scalar $\alpha$ can be omitted in the
computation of $\bz$.  An important feature of this formulation is that
each element of $\bz$ is computed as a multi-linear expression in the 6
image coordinates.  This can be seen from the fact that each point
appears exactly once in the computations of the two determinants in
each element of $\bz$.

This formulation of $\bz$ allows us to rewrite the constraint
as $\bz \cdot \bs$=$\bl_{1} \cdot \by_{1}$= 0 with
\begin{equation}
  \bl_{1} =
  \bl_{26} D_{354} \tI_{1} + \bl_{36} D_{245} \tI_{2} + \bl_{46} D_{253} \tI_{3} +
  \bl_{45} D_{263} \tI_{4} \!+ \bl_{35} D_{246} \tI_{5}
\end{equation}
where $\bl_{ij}$=$\by_{i}$$\times$$\by_{j}$.  $\bl_{1}$ depends on the
five image points $\by_{2}$,...,$\by_{6}$ and on the elements of $\bs$.
A similar exercise can be made for the other five image points and in
general we can write the matching constraint as $\bz \cdot
\bs$=$\bl_{k} \cdot \by_{k}$=$0$ where $\bl_{k}$ depends on $\bs$ and
five image points: $\{ \by_{i}, i$$\neq$$k \}$.  With this description
of the matching constraint it makes sense to interpret $\bl_{k}$ as the
dual homogeneous coordinates of a line in the image plane.  To each of
the 6 image points, $\by_{k}$, there is a corresponding line,
$\bl_{k}$, and the constraint is satisfied if any of the 6 lines
intersects its corresponding image point.
The existence of the lines allows us to quantify the matching
constraint in terms of the Euclidean distance in the image between a
point and its corresponding line.  Assuming that $\by_{k}$ and
$\bl_{k}$ have been suitably normalized, their distance is given simply as
\begin{equation}
  d_{k} = | \by_{k} \cdot \bl_{k} |
  \label{eq:defd}
\end{equation}

\vspace*{-1mm}
\subsection{Estimation of $\bs$}
\label{sec:ests}

$\bs$ can be computed from \eqref{eq:defs}, given that 3D positions are
available, but it can also be estimated from observations of the 6
image points based on the constraint. For
example, from only three observations of the 5-dimensional vector
$\bz$, $\bs$ can be restricted to a 2-dimensional subspace of
$\bbR^{5}$.  From this subspace, $\bs$ can be determined using the
internal constraint \cite{Quan1995PAMI}. This gives in general three
solutions for $\bs$, that satisfy the internal constraint and are
unique except for degenerate cases.  This approach
was used in \cite{NordZogICPR2010}.

Alternatively, for $B \geq 4$ observations of $\bz$ a simple linear
method finds $\bs$ as a total least squares solution of minimizing $\|
\bZ \: \bs \|$ for $\| \bs \|$=$1$, where $\bZ$ is a $B \times 5$
matrix consisting of the observed vectors $\bz$ in its rows. $\bz$ is
then given by the right singular vector of $\bZ$ corresponding to the
smallest singular value. This approach has the advantage of producing a
single solution for $\bs$ which, on the other hand, may not satisfy the
internal constraint. However, this can be compensated for by including
a large number of observations, $B$, in the estimation of $\bs$. This
is the estimation strategy we use in this paper
and it works well, provided that there are enough images in each sequence.

\subsection{Matching score}
\label{sec:score}

In the case of motion segmentation we want to be able to consider a set
of 6 points, estimate $\bs$, and then see how
well this $\bs$ matches to the their trajectories. The
matching between $\bs$ and observations of the 6 points over time is
measured as follows. For each observation (at time $t$) of the 6 points
$\by_{1}(t)$,...,$\by_{6}(t)$ we use $\bs$ to compute the 6
corresponding lines, $\bl_{1}(t)$,...,$\bl_{6}(t)$, and then compute
the distances $d_{k}$ from \eqref{eq:defd}.  Finally, we compute a
matching score $\tilde{E}$ of the 6 point trajectories:
\begin{equation}
  \tilde{E}(P_1, \ldots ,P_6) =
  \underset{t}{\textnormal{median}} \left[ d_{1}^{2}(t) + \ldots + d_{6}^{2}(t) \right]^{1/2},
  \label{eq:score}
\end{equation}
where $P_{k}$ denotes image point $k$, but without reference to a
particular image position in a particular frame.  The median operation
is used here in order to effectively reduce the influence of possible
outliers.

\begin{algorithm}[t]
\scriptsize 
\DontPrintSemicolon Create spatial clusters using k-means\;
\ForEach{point $P_k$} {

\ForEach{cluster $C_j$} {
  Select 6 points $\{P_k,P_2^j,...,P_6^j\}$. \;
  Calculate score $E(P_k,C_j)$ from (\ref{eq:score}). \;
}
 Assign $P_i$ to cluster with $\min(E(P_k,C_j))$. \;
} Reject inconsistent clusters. \; Initial NBC merging. \; Final
refinement merging. \; \caption{Motion segmentation pseudocode.}
\label{alg:algorithm}
\end{algorithm}

%% file: algorithm.tex
\section{A motion segmentation algorithm}
\label{sec:algo}

In this section we describe a simple yet effective algorithm that can
be used for the segmentation of multiple moving rigid 3D objects in a
scene. The input data is the number of motion segments and a set of $N$
point trajectories over a set of images in an image sequence. Our
approach includes: a \emph{spatial initialisation} step for
establishing the initial motion hypotheses (or \emph{seed} clusters),
from which the segmentation will evolve; a \emph{classification} stage,
whereby each tracked point $P_{k}$, is assigned to the appropriate motion
cluster; and a \emph{merging} step, that combines clusters based on
their similarity, to form the final number of moving objects in the
scene.

\begin{figure}[tb]
	\centering
	 \includegraphics[width=0.32\textwidth]{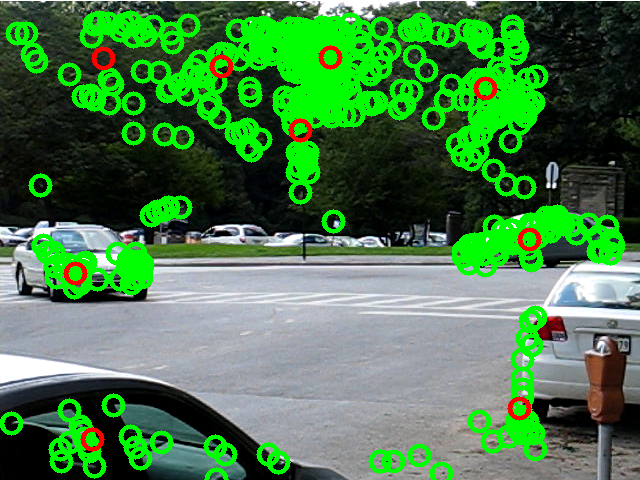}
   \includegraphics[width=0.32\textwidth]{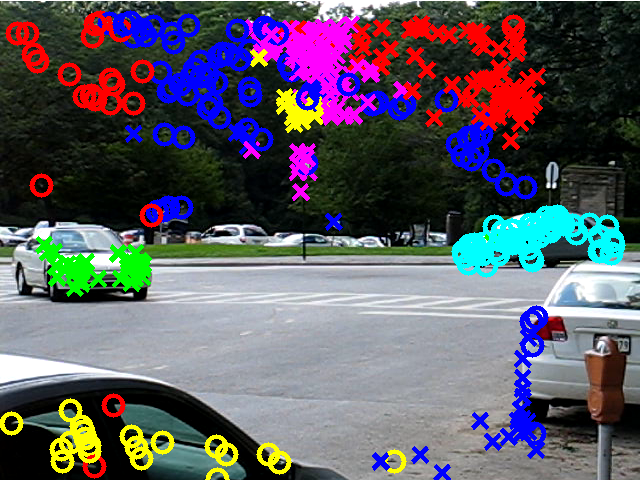}
   \includegraphics[width=0.32\textwidth]{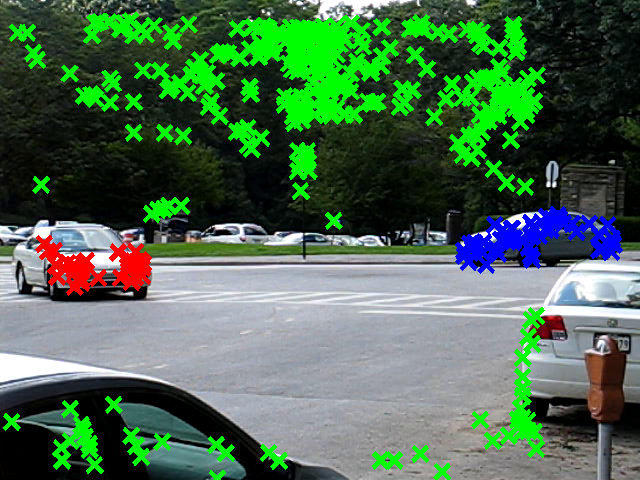}
	\caption{A K-means initialisation example on the left. On the centre the classification result before the merging, and the final merged results on the left.}
	\label{fig:Example}
\end{figure}

\textbf{Initialisation:} \label{sec:init}
The first step is the generation of initial 6-point
clusters, each representing a 3D motion hypothesis.  
For this we use spatial K-means clustering in the image domain (see Fig. \ref{fig:Example}). The initial clustering is carried out in an arbitrary frame
from each sequence (usually the first or the last). We define a seed
cluster $C_j$=$\{P_1^j$,...,$P_I^j\}$ as the $I$ points at minimum
distance to each K-means center.   From the subsequent
computations it is required that $I$$\geq$5, and we use $I$=6.

\textbf{Point classification:} \label{sec:classification}
Following the initialisation step, we assign the remaining points to
the appropriate seed cluster. For each of the unclassified points $P_k$
and for each seed cluster $C_j$, we estimate $\bs$ according to Sec.
\ref{sec:ests} and compute a point-to-cluster score $E$ from
\eqref{eq:score} as $E(P_k,C_j)$=$\tilde{E}(P_k,P_2^j$,...,$P_6^j)$. This
gives $M(N$-$6M)$ score calculations in total, and produces an $M
\times (N$-$6M)$ matrix $\bA$=$[ a_{ik} ]$ , with column $k$ referring
to particular point $P_{k}$ and element $a_{ik}$ as the index of the
cluster that has the $i$-th smallest score relative to $P_{k}$.  We
employ a ``winner takes all'' approach with $P_k$ assigned to the
cluster that produces the lowest score, i.e., to the cluster index
$a_{1k}$.  This implies that the clusters will grow during the
classification step, however, it should be noted that the scores for a
particular point are always computed relative to the seed clusters.
Note also that there is no threshold associated with the actual
classification stage. A typical classification result can be seen in
Fig. \ref{fig:Example}. The growth of the clusters is independent of the
order that the points are classified, so the latter may be considered in parallel,
leading to a very efficient and fast implementation.

\textbf{Cluster merging and rejection}\label{sec:merging}
This is the final stage of our method, and results in the generation of
motion consistent clusters each associated with a unique moving object
in the scene. This stage consists of a quick \emph{cluster rejection}
step; an \emph{initial merging} step using redundant classification
information; and a final merging or \emph{refinement} step where
intermediate clusters are combined using agglomerative clustering based on some similarity measure.

-\textit{Cluster rejection:}
Any clusters that contain very few points (e.g. $\leq$7) are indicative of seed initialisation between motion boundaries, and 
represent unique and erroneous motion hypotheses. Therefore, any such clusters are promptly removed and their points
re-classified with the remaining clusters.

-\textit{Initial merging:}
A direct result of the classification in Sec. \ref{sec:classification}
is the matrix $\bA$, where so far we have only used the top row in
order to classify points.  However, $\bA$ provides also information on
cluster similarity, which we can exploit to infer initial merge
pairings.  We call this ``Next-Best Classification'' (NBC) merging and
we now look at the cluster with the second best score for each
point, since it contains enough discriminative power to accurately
merge clusters.
NBC merging involves generating the zero-diagonal sparse symmetric $M$$\times$$M$
matrix $\bL$=$[l_{ij}]$ that contains the merging similarity
between the clusters.  Its elements are defined as:
\begin{equation}
  l_{ij} =
  \sum_{k = 1}^{N - 6 M} \left[ \frac{1(k, i, j)}{E(P_k,C_j)} + \frac{1(k, j, i)}{E(P_k,C_i)} \right],
  \label{eq:L_matrix}
\end{equation}
where the summation is made over the $N - 6 M$ points not included in
the seed clusters.  $1(k, i, j)$ is an indicator function that takes
the value 1 when $a_{1k}$=$i$ and $a_{2k}$=$j$ and 0 otherwise.  In other
words, this function is =$1$ iff $P_{k}$ is assigned to cluster $i$ and
has cluster $j$ as second best option.

The matrix $\bL$ describes all the consistent pairings inferred by the NBC merging. However, since
usually inconsistent clusters will generate non-zero entries in $\bL$
we need to threshold out low response entries due to noise.  Using a
threshold $\tau$ we obtain the sparser \textit{adjacency matrix}
$\bL^*$. From $\bL^{\star}$ we can then construct an
undirected graph $G$ which contains the intermediate clusters as
disconnected sub-graphs.
If $\bL^*$ is insufficient to provide the final motion clusters, due to for example noisy data, then
a final refinement step may be
required. The result of the cluster rejection and initial
merging steps is a set of $\tilde{M}$$\leq$$M$ clusters $\tilde{C}_{1}$,...,$\tilde{C}_{\tilde{M}}$.

-\textit{Refinement merging:}
The last step involves the merging of the
intermediate clusters, (resulting from the NBC merging), into the final clusters each representing a distinct motion hypothesis. This
is achieved by pairwise agglomerative clustering and a maximum similarity
measure between clusters. 
Assume that we wish to merge two clusters, say $\tilde{C}_1$ and
$\tilde{C}_2$. We can generate $K$ 6-point mixture clusters
$\tilde{C}'$ by randomly selecting 3 points each from $\tilde{C}_1$ and
$\tilde{C}_2$.  If $\tilde{C}_1$ and $\tilde{C}_2$ belong to the same
motion-consistent object and there is little noise present, we expect
the scores $\tilde{E}$ calculated for each selection of $C'$ to be
grouped near zero, with little variation and few outliers. Conversely, if $\tilde{C}_1$ and
$\tilde{C}_2$ come from different objects, $\tilde{E}$ should exhibit a
larger dispersion and be grouped further away from zero.
Instead of defining the similarity based on location and dispersion of
sample statistics, we fit a parametric model to the sample data (using
Maximum Likelihood Estimation) and compute the statistics from the
model parameters. This allows for a much smaller number of samples and
a more accurate estimate than what can be obtained from sample
statistics (e.g. mean and variance). Given therefore that the scores in
(\ref{eq:score}) should generally group around a median value with a
few extremal outliers and assuming that the distances $d_{k}$ in (\ref{eq:defd})
are i.i.d., then the score distribution may be well approximated
by a Generalised Extreme Value (GEV) distribution \cite{Lead1983}. 
A robust indication of average location in a data sample with outliers
is the mode, which for the GEV model can be computed by:
\begin{equation}
	\widetilde{m} =
    \mu+\sigma\left[{(1+\xi)^{-\xi}-1}\right]/{\xi} \ \ \ \textnormal{for} \ \ \xi \neq 0, \label{eq:GEV_mode}
\end{equation}
where $\mu$, $\sigma$ and $\xi$ are the location, scale and shape
parameters respectively recovered by the MLE. Using this as a similarity
metric we can merge two clusters when (\ref{eq:GEV_mode}) is small or
reject them when it is large. 
The clustering proceeds until we reach the pre-defined number of motions in the scene.
The overall method is included
in pseudocode in Algorithm \ref{alg:algorithm}.

\begin{table}[t]
\begin{centering}
\begin{tabular}{|l|c|c|c|c|c|c|c|c|c|c|c|c|c|c|c|}
\hline
 & {\tiny GPCA} & {\tiny LSA} & {\tiny RANSAC} & {\tiny MSL} & {\tiny ALC} & {\tiny SSC} &{\tiny SCC} & {\tiny SPC} & {\tiny SC} & {\tiny LP} & {\tiny NNMF} & {\tiny NSI} & {\tiny LLMC} & {\tiny CCS} & \textbf{\tiny MSPC}\tabularnewline
\hline
\multicolumn{16}{l}{\emph{\tiny Checkerboard: 78 sequences}}\tabularnewline
\hline
{\tiny Mean:} & {\tiny 6.09} & {\tiny 2.57} & {\tiny 6.52} & {\tiny 4.46} & {\tiny 1.55} & {\tiny 1.12} & {\tiny 1.77} & {\tiny 4.49} & {\tiny 0.85} & {\tiny 3.21} & {\tiny -} & {\tiny 3.75} & {\tiny 4.37} & {\tiny 16.37} & \textbf{\tiny 0.41}\tabularnewline
\hline
{\tiny Median:} & {\tiny 1.03} & {\tiny 0.27} & {\tiny 1.75} & {\tiny 0.00} & {\tiny 0.29} & {\tiny 0.00} & {\tiny 0.00} & {\tiny 3.69} & {\tiny 0.00} & {\tiny 0.11} & {\tiny -} & {\tiny -} & {\tiny 0.00} & {\tiny 10.62} & {\tiny 0.00}\tabularnewline
\hline
\multicolumn{16}{l}{\emph{\tiny Traffic: 31 sequences}}\tabularnewline
\hline
{\tiny Mean:} & {\tiny 1.41} & {\tiny 5.43} & {\tiny 2.55} & {\tiny 2.23} & {\tiny 1.59} & \textbf{\tiny 0.02} & {\tiny 0.63} & {\tiny 0.22} & {\tiny 0.90} & {\tiny 0.33} & {\tiny 0.1-} & {\tiny 1.69} & {\tiny 0.84} & {\tiny 5.27} & {\tiny 0.09}\tabularnewline
\hline
{\tiny Median:} & {\tiny 0.00} & {\tiny 1.48} & {\tiny 0.21} & {\tiny 0.00} & {\tiny 1.17} & {\tiny 0.00} & {\tiny 0.14} & {\tiny 0.00} & {\tiny 0.00} & {\tiny 0.00} & {\tiny 0.--} & {\tiny -} & {\tiny 0.00} & {\tiny 0.00} & {\tiny 0.00}\tabularnewline
\hline
\multicolumn{16}{l}{\emph{\tiny Articulated: 11 sequences}}\tabularnewline
\hline
{\tiny Mean:} & {\tiny 2.88} & {\tiny 4.10} & {\tiny 7.25} & {\tiny 7.23} & {\tiny 10.70} & \textbf{\tiny 0.62} & {\tiny 4.02} & {\tiny 2.18} & {\tiny 1.71} & {\tiny 4.06} & {\tiny 10.--} & {\tiny 8.05} & {\tiny 6.16} & {\tiny 17.58} & {\tiny 0.95}\tabularnewline
\hline
{\tiny Median:} & {\tiny 0.00} & {\tiny 1.22} & {\tiny 2.64} & {\tiny 0.00} & {\tiny 0.95} & {\tiny 0.00} & {\tiny 2.13} & {\tiny 0.00} & {\tiny 0.00} & {\tiny 0.00} & {\tiny 2.6-} & {\tiny -} & {\tiny 1.37} & {\tiny 7.07} & {\tiny 0.00}\tabularnewline
\hline
\multicolumn{16}{l}{\emph{\tiny All: 120 sequences}}\tabularnewline
\hline
{\tiny Mean:} & {\tiny 4.59} & {\tiny 3.45} & {\tiny 5.56} & {\tiny 4.14} & {\tiny 2.40} & {\tiny 0.82} & {\tiny 1.68} & {\tiny 3.18} & {\tiny 0.94} & {\tiny 2.20} & {\tiny -} & {\tiny -} & {\tiny 3.62} & {\tiny 12.16} & \textbf{\tiny 0.37}\tabularnewline
\hline
{\tiny Median:} & {\tiny 0.38} & {\tiny 0.59} & {\tiny 1.18} & {\tiny 0.00} & {\tiny 0.43} & {\tiny 0.00} & {\tiny 0.07} & {\tiny 1.08} & {\tiny 0.00} & {\tiny 0.00} & {\tiny -} & {\tiny -} & {\tiny 0.00} & {\tiny 0.00} & {\tiny 0.00}\tabularnewline
\hline
\end{tabular}
\par\end{centering}\caption{2 motion results} \label{tab:2motions}

\end{table}

%% file: experiments.tex
\section{Experimental results} \label{sec:experiments}

We have carried out experiments on real image sequences from the
Hopkins~155 database \cite{TronVidal2007CVPR}. It includes motion
sequences of 2 and 3 objects, of various degrees of classification
difficulty and is corrupted by tracking noise, but without any missing entries or
outliers.
Typicall parameter settings for these experiments were: $M$=10-40
K-means clusters at the first or last frame of the sequence, reject
clusters of $\leq$7 points, and $K$=50-100 mixture samples for the
final merge (where necessary). Our results for 2 and 3 motions and the
whole database are presented and compared with other state-of-the-art
and baseline methods in Tables \ref{tab:2motions}--\ref{tab:Allmotions}. 

Our approach (Multiple Six
Point Consistency - MSPC) outperforms every other method in the
literature overall, in 2 and 3 motions and for all sequences combined.
We achieve an overall classification error of 0.37\% for two motions,
less than 1/2 than the best reported result (SSC); an overall error of
1.32\% for three motions, about 2/3 of the best reported result (SC);
and an overall error of  0.59\% for the whole database, less than 1/2
than the best reported result (SC). We also come first for the
checkerboard sequences constituting the majority of the data, with
almost 1/2 the classification errors reported by the SC method. For the
articulated and traffic sequences (which are problematic for most
methods) we perform well, coming a very close second to the best
performing SSC or NNMF.

\begin{figure}
	\centering
	 \includegraphics[width=0.35\textwidth]{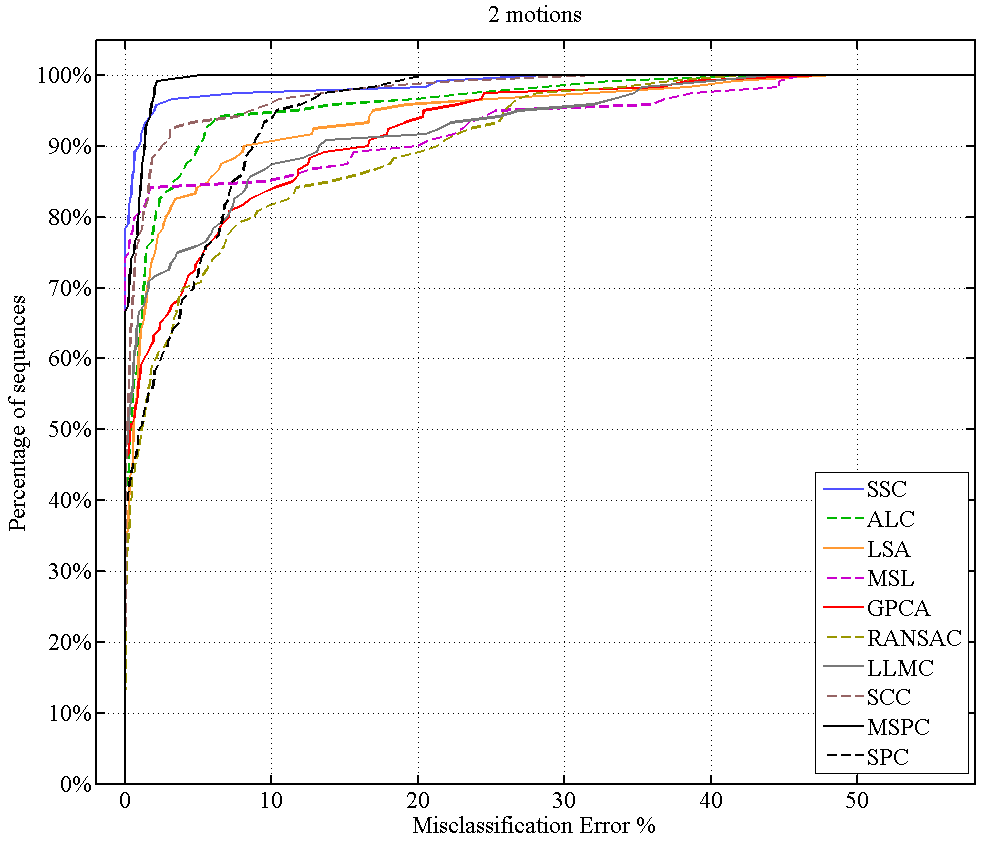}
     \includegraphics[width=0.35\textwidth]{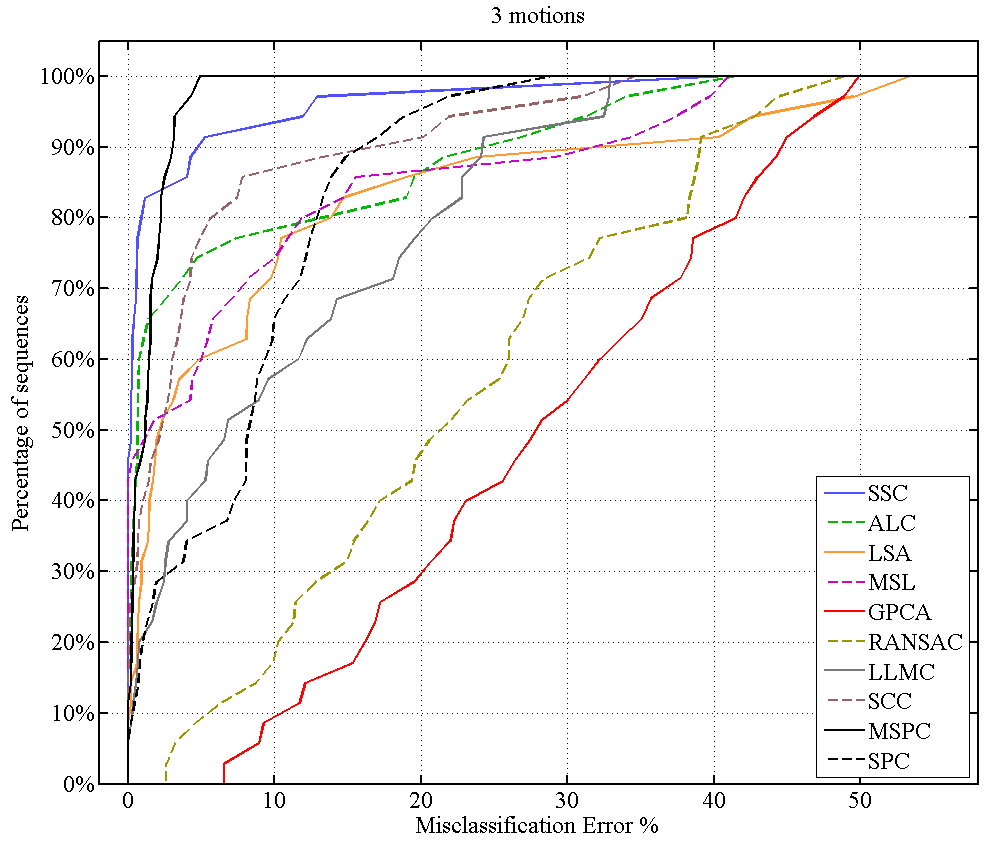}
	\caption{Cumulative distributions of the errors per sequence for two and three motions.}
	\label{fig:CDF}
\end{figure}

\begin{figure}
	\centering
	 \includegraphics[width=0.35\textwidth]{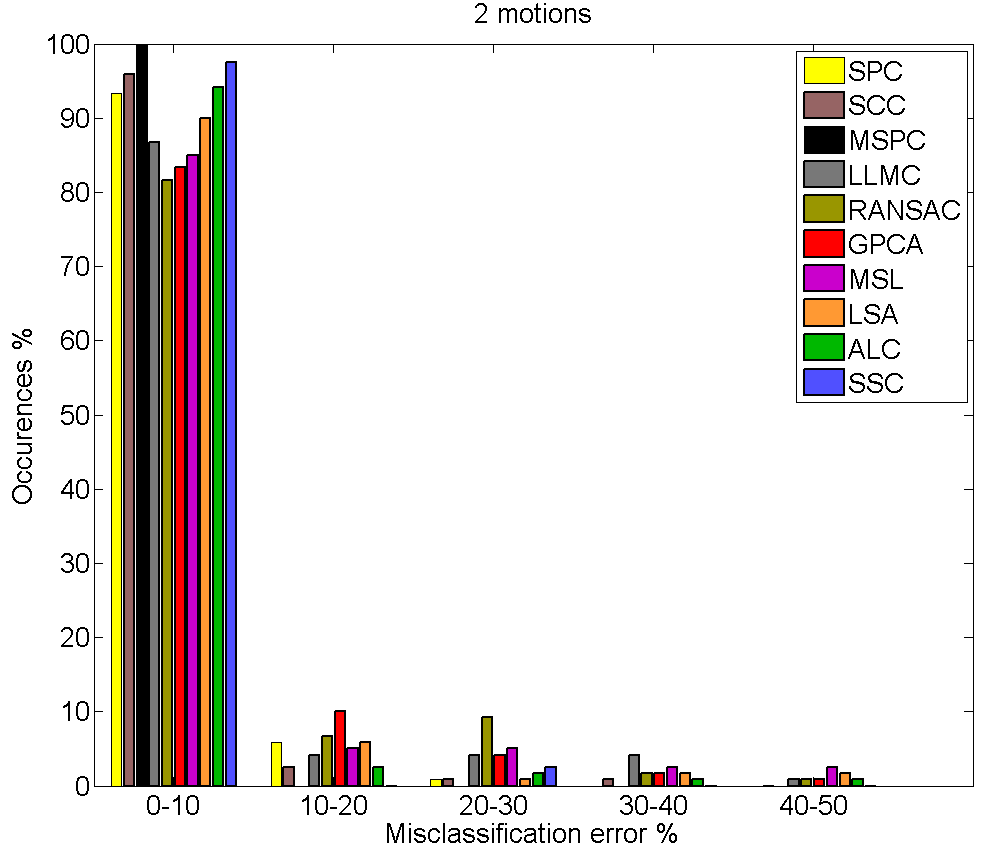}
   \includegraphics[width=0.35\textwidth]{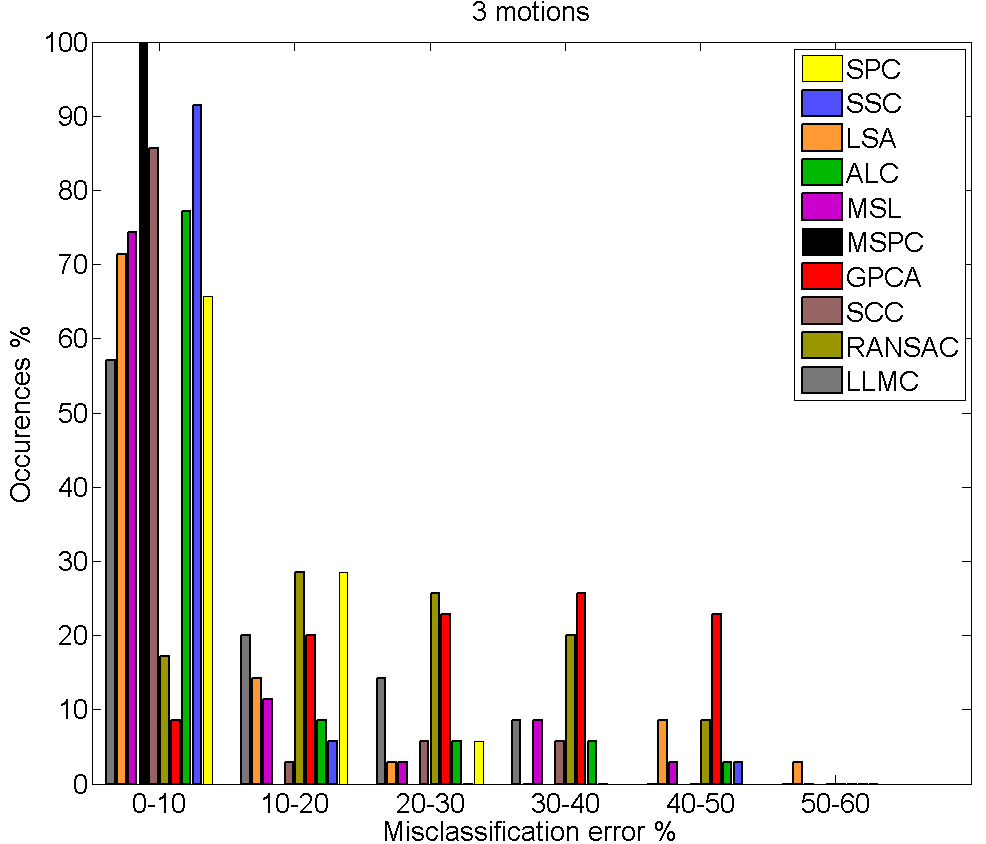}
	\caption{Histograms of the errors per sequence for two and three motions.}
	\label{fig:Hist}
\end{figure}

From the cumulative distributions in Fig. \ref{fig:CDF} we see that our
method outperforms all others (where available) with only the SSC being
slightly better (between 0.5-1\% error) for 20-30\% of the sequences.
However, SSC soon degrades quite rapidly for the remaining 5-20\% of
the data with an error differential between 15-35\% relative to MSPC.
Furthermore, our method degrades gracefully from 2 to 3 motions as we
do not have misclassification errors greater than 5\% for any of the
sequences, unlike SSC which produces a few errors between 10-20\% and
40-50\%.  This is better illustrated in the histograms in Fig.
\ref{fig:Hist}.

\begin{table}[t]
\begin{centering}
\begin{tabular}{|l|c|c|c|c|c|c|c|c|c|c|c|c|c|c|c|}
\hline
 & {\tiny GPCA} & {\tiny LSA} & {\tiny RANSAC} & {\tiny MSL} & {\tiny ALC} & {\tiny SSC} &{\tiny SCC} & {\tiny SPC} & {\tiny SC} & {\tiny LP} & {\tiny NNMF} & {\tiny NSI} & {\tiny LLMC} & {\tiny CCS} & \textbf{\tiny MSPC}\tabularnewline
\hline
\multicolumn{16}{l}{\emph{\tiny Checkerboard: 26 sequences}}\tabularnewline
\hline
{\tiny Mean:} & {\tiny 31.95} & {\tiny 5.80} & {\tiny 25.78} & {\tiny 10.38} & {\tiny 5.20} & {\tiny 2.97} & {\tiny 6.23} & {\tiny 10.71} & {\tiny 2.15} & {\tiny 8.34} & {\tiny -} & {\tiny 2.92} & {\tiny 10.70} & {\tiny 28.63} & \textbf{\tiny 1.43}\tabularnewline
\hline
{\tiny Median:} & {\tiny 32.93} & {\tiny 1.77} & {\tiny 26.01} & {\tiny 4.61} & {\tiny 0.67} & {\tiny 0.27} & {\tiny 1.70} & {\tiny 9.61} & {\tiny 0.47} & {\tiny 5.35} & {\tiny -} & {\tiny -} & {\tiny 9.21} & {\tiny 33.21} & {\tiny 1.25}\tabularnewline
\hline
\multicolumn{16}{l}{\emph{\tiny Traffic: 7 sequences}}\tabularnewline
\hline
{\tiny Mean:} & {\tiny 19.83} & {\tiny 25.07} & {\tiny 12.83} & {\tiny 1.80} & {\tiny 7.75} & {\tiny 0.58} & {\tiny 1.11} & {\tiny 0.73} & {\tiny 1.35} & {\tiny 2.34} & \textbf{\tiny 0.1-} & {\tiny 1.67} & {\tiny 2.91} & {\tiny 3.02} & {\tiny 0.71}\tabularnewline
\hline
{\tiny Median:} & {\tiny 19.55} & {\tiny 23.79} & {\tiny 11.45} & {\tiny 0.00} & {\tiny 0.49} & {\tiny 0.00} & {\tiny 1.40} & {\tiny 0.73} & {\tiny 0.19} & {\tiny 0.19} & {\tiny 0.--} & {\tiny -} & {\tiny 0.00} & {\tiny 0.18} & {\tiny 0.36}\tabularnewline
\hline
\multicolumn{16}{l}{\emph{\tiny Articulated: 2 sequences}}\tabularnewline
\hline
{\tiny Mean:} & {\tiny 16.85} & {\tiny 7.25} & {\tiny 21.38} & {\tiny 2.71} & {\tiny 21.08} & \textbf{\tiny 1.42} & {\tiny 5.41} & {\tiny 6.91} & {\tiny 4.26} & {\tiny 8.51} & {\tiny 15.--} & {\tiny 6.38} & {\tiny 5.60} & {\tiny 44.89} & {\tiny 2.13}\tabularnewline
\hline
{\tiny Median:} & {\tiny 28.66} & {\tiny 7.25} & {\tiny 21.38} & {\tiny 2.71} & {\tiny 21.08} & {\tiny 0.00} & {\tiny 5.41} & {\tiny 6.91} & {\tiny 4.26} & {\tiny 8.51} & {\tiny 15.--} & {\tiny -} & {\tiny 5.60} & {\tiny 44.89} & {\tiny 2.13}\tabularnewline
\hline
\multicolumn{16}{l}{\emph{\tiny All: 35 sequences}}\tabularnewline
\hline
{\tiny Mean:} & {\tiny 28.66} & {\tiny 9.73} & {\tiny 22.94} & {\tiny 8.23} & {\tiny 6.69} & {\tiny 2.45} & {\tiny 5.16} & {\tiny 8.49} & {\tiny 2.11} & {\tiny 7.66} & {\tiny -} & {\tiny -} & {\tiny 8.85} & {\tiny 26.18} & \textbf{\tiny 1.32}\tabularnewline
\hline
{\tiny Median:} & {\tiny 28.26} & {\tiny 2.33} & {\tiny 22.03} & {\tiny 1.76} & {\tiny 0.67} & {\tiny 0.20} & {\tiny 1.58} & {\tiny 8.36} & {\tiny 0.37} & {\tiny 5.60} & {\tiny -} & {\tiny -} & {\tiny 3.19} & {\tiny 31.74} & {\tiny 1.17}\tabularnewline
\hline
\end{tabular}
\par\end{centering}\caption{3 motion results} \label{tab:3motions}

\end{table}
\vspace*{-1mm}
\begin{table}[t]

\begin{centering}
\begin{tabular}{|l|c|c|c|c|c|c|c|c|c|c|c|c|c|c|c|}
\hline
 & {\tiny GPCA} & {\tiny LSA} & {\tiny RANSAC} & {\tiny MSL} & {\tiny ALC} & {\tiny SSC} &{\tiny SCC} & {\tiny SPC} & {\tiny SC} & {\tiny LP} & {\tiny NNMF} & {\tiny NSI} & {\tiny LLMC} & {\tiny CCS} & \textbf{\tiny MSPC}\tabularnewline
\hline
\multicolumn{16}{l}{\emph{\tiny Checkerboard: 104 sequences}}\tabularnewline
\hline
{\tiny Mean:} & \emph{\tiny 12.55} & \emph{\tiny 3.37} & \emph{\tiny 11.33} & \emph{\tiny 5.94} & {\tiny 2.47} & \emph{\tiny 1.58} & \emph{\tiny 2.88} & {\tiny 6.05} & {\tiny 1.17} & \emph{\tiny 4.49} & {\tiny -} & \emph{\tiny 3.54} & \emph{\tiny 5.95} & \emph{\tiny 19.43} & \textbf{\tiny 0.66}\tabularnewline
\hline
{\tiny Median:} & {\tiny -} & {\tiny -} & {\tiny -} & {\tiny -} & {\tiny 0.31} & {\tiny -} & {\tiny -} & {\tiny 5.27} & {\tiny 0.00} & {\tiny -} & {\tiny -} & {\tiny -} & {\tiny -} & {\tiny -} & {\tiny 0.25}\tabularnewline
\hline
\multicolumn{16}{l}{\emph{\tiny Traffic: 38 sequences}}\tabularnewline
\hline
{\tiny Mean:} & \emph{\tiny 4.80} & \emph{\tiny 9.04} & \emph{\tiny 4.44} & \emph{\tiny 2.15} & {\tiny 2.77} & \textbf{\emph{\tiny 0.12}} & \emph{\tiny 0.71} & {\tiny 0.31} & {\tiny 0.98} & \emph{\tiny 0.70} & \textbf{\emph{\tiny 0.1-}} & \emph{\tiny 1.68} & \emph{\tiny 1.22} & \emph{\tiny 4.85} & {\tiny 0.20}\tabularnewline
\hline
{\tiny Median:} & {\tiny -} & {\tiny -} & {\tiny -} & {\tiny -} & {\tiny 1.10} & {\tiny -} & {\tiny -} & {\tiny 0.00} & {\tiny 0.00} & {\tiny -} & {\tiny -} & {\tiny -} & {\tiny -} & {\tiny -} & {\tiny 0.00}\tabularnewline
\hline
\multicolumn{16}{l}{\emph{\tiny Articulated: 13 sequences}}\tabularnewline
\hline
{\tiny Mean:} & \emph{\tiny 5.02} & \emph{\tiny 4.58} & \emph{\tiny 9.42} & \emph{\tiny 6.53} & {\tiny 13.71} & \textbf{\emph{\tiny 0.74}} & \emph{\tiny 4.23} & {\tiny 2.91} & {\tiny 2.10} & \emph{\tiny 4.74} & \emph{\tiny 10.76} & \emph{\tiny 7.79} & \emph{\tiny 6.07} & \emph{\tiny 21.78} & {\tiny 1.13}\tabularnewline
\hline
{\tiny Median:} & {\tiny -} & {\tiny -} & {\tiny -} & {\tiny -} & {\tiny 3.46} & {\tiny -} & {\tiny -} & {\tiny 0.00} & {\tiny 0.00} & {\tiny -} & {\tiny -} & {\tiny -} & {\tiny -} & {\tiny -} & {\tiny 0.00}\tabularnewline
\hline
\multicolumn{16}{l}{\emph{\tiny All: 155 sequences}}\tabularnewline
\hline
{\tiny Mean:} & {\tiny 10.34} & {\tiny 4.94} & {\tiny 9.76} & {\tiny 5.03} & {\tiny 3.56} & {\tiny 1.24} & \emph{\tiny 2.46} & {\tiny 4.38} & {\tiny 1.20} & \emph{\tiny 3.43} & {\tiny -} & {\tiny -} & \emph{\tiny 4.8} & \emph{\tiny 15.32} & \textbf{\tiny 0.59}\tabularnewline
\hline
{\tiny Median:} & {\tiny 2.54} & {\tiny 0.90} & {\tiny 3.21} & {\tiny 0.00} & {\tiny 0.50} & {\tiny 0.00} & {\tiny -} & {\tiny 1.95} & {\tiny 0.00} & {\tiny -} & {\tiny -} & {\tiny -} & {\tiny -} & {\tiny -} & {\tiny 0.00}\tabularnewline
\hline
\end{tabular}
\par\end{centering}
\caption{All motion results (italics are approximated from Tables \ref{tab:2motions} and \ref{tab:3motions})} \label{tab:Allmotions}

\end{table}
\vspace*{-1mm}

%% file: conclusion.tex
\section{Conclusion}\label{sec:conclusion}

We have presented a method for segmenting moving objects using the geometry of
6 points to infer motion consistency.  Our evaluations on the
Hopkins~155 database have shown superior results than current state-of-the-art methods,
both in terms of overall performance and in terms of maximum errors.
The method finds initial cluster seeds in the spatial domain, and then
classifies points as belonging to the cluster that minimizes a motion
consistency score. The score is based on a geometric matching error
measured in the image, implicitly describing how consistent the motion
trajectories of 6 points are relative to a rigid 3D motion. Finally,
the resulting clusters are merged by agglomerative clustering using a
similarity criterion.